\documentclass[10.7pt,]{article}

\usepackage[letterpaper, margin=2.54cm, top=2.54cm]{geometry}
\usepackage[super,comma,sort&compress]{natbib}
\usepackage{lmodern}
\usepackage{authblk} % To add affiliations to authors
\usepackage{amssymb,amsmath}
\usepackage{wrapfig}
\usepackage{graphicx,grffile}
\usepackage[labelfont=bf,labelsep=period]{caption}
\usepackage{ifxetex,ifluatex}
\usepackage{fixltx2e} % provides \textsubscript
\usepackage{booktabs}
\usepackage{multirow}
\usepackage{graphicx}
\usepackage{booktabs}
\usepackage{graphicx}
\usepackage{tablefootnote}
\usepackage[table,xcdraw]{xcolor}
\ifnum 0\ifxetex 1\fi\ifluatex 1\fi=0 % if pdftex
  \usepackage[T1]{fontenc}
  \usepackage[utf8]{inputenc}
\else % if luatex or xelatex
  \ifxetex
    \usepackage{mathspec}
  \else
    \usepackage{fontspec}
  \fi
  \defaultfontfeatures{Ligatures=TeX,Scale=MatchLowercase}
\fi
\IfFileExists{upquote.sty}{\usepackage{upquote}}{}
\IfFileExists{microtype.sty}{%
	\usepackage{microtype}
	\UseMicrotypeSet[protrusion]{basicmath} % disable protrusion for tt fonts
}{}

\usepackage{sectsty}
\allsectionsfont{\fontsize{10}{10}\selectfont}
\subsectionfont{\itshape\bfseries\fontsize{10}{10}\selectfont}
\subsubsectionfont{\normalfont\itshape}

\makeatletter
\renewcommand\subsubsection{\@startsection{subsubsection}{3}{\z@}%
	{-3.25ex\@plus -1ex \@minus -.2ex}%
    {-1.5ex \@plus -.2ex}% Formerly 1.5ex \@plus .2ex
    {\normalfont\itshape}}
\makeatother

\makeatletter % Reference list option change
\renewcommand\@biblabel[1]{#1.} % from [1] to 1
\makeatother %

\usepackage{etoolbox}
\makeatletter
\patchcmd{\@maketitle}{\LARGE}{\bfseries\fontsize{15}{16}\selectfont}{}{}
\makeatother

\makeatletter
\def\maxwidth{\ifdim\Gin@nat@width>\linewidth\linewidth\else\Gin@nat@width\fi}
\def\maxheight{\ifdim\Gin@nat@height>\textheight\textheight\else\Gin@nat@height\fi}
\makeatother

\setkeys{Gin}{width=\maxwidth,height=\maxheight,keepaspectratio}
\ifx\paragraph\undefined\else
\let\oldparagraph\paragraph
\renewcommand{\paragraph}[1]{\oldparagraph{#1}\mbox{}}
\fi
\ifx\subparagraph\undefined\else
\let\oldsubparagraph\subparagraph
\renewcommand{\subparagraph}[1]{\oldsubparagraph{#1}\mbox{}}
\fi
\usepackage{hyperref}
\hypersetup{
	unicode=true,
	pdftitle={My Cool Title Here},
	pdfauthor={Author One, Author Two, Author Three},
	pdfkeywords={keyword1, keyword2},
	pdfborder={0 0 0},
	breaklinks=true
}
\title{\vspace{-2em} Crowdsourcing with Enhanced Data Quality Assurance: An Efficient Approach to Mitigate Resource Scarcity Challenges in Training Large Language Models for Healthcare}
\author[ ]{\bf\fontsize{13}{14}\selectfont Prosanta Barai, MS\textsuperscript{1}, Gondy Leroy,
PhD\textsuperscript{1}, Prakash Bisht\textsuperscript{1}, Joshua M. Rothman, MD \textsuperscript{2}, Sumi Lee, MS \textsuperscript{1}, Jennifer Andrews, PhD \textsuperscript{1}, Sydney A. Rice, MD, MSc \textsuperscript{1}, Arif Ahmed, MS\textsuperscript{1}\vspace{-.7em}}
\affil[1]{\bf\fontsize{13}{14}\selectfont The University of Arizona, Tucson 85721, U.S.A;}
\affil[2]{\bf\fontsize{13}{14}\selectfont UC San Diego Division of Academic General Pediatrics, USA}
\date{} % add no date (by default date is added)

\begin{document}
\maketitle
\vspace{-4em} %separation between the affiliations and abstract
%==============================

%==============================
\section{Abstract}\label{abstract}
%==============================
Large Language Models (LLMs) have demonstrated immense potential in artificial intelligence across various domains, including healthcare. However, their efficacy is hindered by the need for high-quality labeled data, which is often expensive and time-consuming to create, particularly in low-resource domains like healthcare. To address these challenges, we propose a crowdsourcing (CS) framework enriched with quality control measures at the pre-, real-time-, and post-data gathering stages. Our study evaluated the effectiveness of enhancing data quality through its impact on LLMs (Bio-BERT) for predicting autism-related symptoms. The results show that real-time quality control improves data quality by 19\% compared to pre-quality control. Fine-tuning Bio-BERT using crowdsourced data generally increased recall compared to the Bio-BERT baseline but lowered precision. Our findings highlighted the potential of crowdsourcing and quality control in resource-constrained environments and offered insights into optimizing healthcare LLMs for informed decision-making and improved patient care.

%==============================
\section{Introduction}\label{introduction}
%==============================

With recent advancements in artificial intelligence (AI) and machine learning (ML), Large Language Models (LLMs) have emerged as powerful tools that can tap into unprecedented capabilities in natural language understanding (identifying fake news, conversational response generation) \cite{ zellers2019defending, zhang2020dialogpt} and healthcare decision support systems (medical question answering, summarizing electronic health records, pathology text mining) \cite{ thirunavukarasu2023large,  santos2022pathologybert, singhal2023towards, adams2023leveraging}. However, the remarkable performance and efficiency of LLMs often face a challenge: the scarcity of resources required for training these models effectively \cite{masud2012facing}. Models like GPT-3 (175B parameters, 45TB training data) \cite{brown2020language} and BARD (540B parameters, 80B tokens)\cite{ chowdhery2022palm} rely on large corpora of text data. While they provide reasonable out-of-the-box performance, they often require fine-tuning. The required data for training or tuning can be cost-prohibitive or practically impossible in fields such as healthcare, where securing balanced labeled training data remains a significant hurdle \cite{ aydin2019medical, ondov2022survey}. This results in challenges like class imbalance, spurious correlation, and data bias.  Moreover, experts warn that we may soon run out of new text for training due to the low projected availability of unlabeled text data and the memorizing effect of LLM\cite{ villalobos2022will}. However, healthcare practitioners require highly sophisticated and well-trained models, underscoring the urgency of addressing this challenge \cite{thirunavukarasu2023large}.

\subsection{Algorithmic Solutions for ML in Low Resource Domains}
To address the resource limitations, researchers across various domains have proposed various solutions. For example, transfer learning (TL) \cite{pan2010survey}, as introduced by Bozinovski and Fulgosi (1976) \cite{bozinovski1976influence}, involves leveraging the knowledge of a pretrained model from a richer domain to boost the performance of a further downstream task. Even though TL has been useful in several low-resource domain scenarios \cite{peng2019transfer,abad2020cross}, challenges like domain mismatch \cite{niu2020decade}, catastrophic forgetting \cite{kirkpatrick2017overcoming}, and negative transfer \cite{rosenstein2005transfer} persist. If we find a reasonable match between the source and target domain, the computational and data resources necessary to tune billions of parameters are challenging, making it an undesirable choice to adapt \cite{brown2020language}. Another alternative well-established technique is resampling \cite{estabrooks2004multiple}, as documented in Estabrooks’s work from 2004. This method, characterized as an external approach, allows for the oversampling of minority classes or the undersampling of majority classes \cite{chawla2002smote}. However, it comes with several associated challenges, including reliance on classification type \cite{varotto2021comparison} and the potential for information loss attributable to data redundancy \cite{susan2021balancing}. Another noteworthy approach to the low resource domain problem is data augmentation, which includes synthetic data augmentation. (SDA) \cite{kumar2020data} and crowdsourcing \cite{doan2011crowdsourcing}. In SDA, data can be augmented using either data warping \cite{wong2016understanding} (transformations applied in the data space) or a pretrained generative model \cite{kumar2020data}. Literature shows improvement in model performance where the amount of real data bounds improvement \cite{wong2016understanding}. Additionally, issues such as target domain representation \cite{dou2019unsupervised} and self-consuming generative process \cite{alemohammad2023self} continue to challenge the effectiveness of SDA.

\subsection{Crowdsourcing Solutions for ML in Low Resource Domains}

With crowdsourcing, humans, usually laypersons and not domain experts, are recruited to generate data. Domain relevance (alignment of a crowd worker’s task with a field of interest) and creating diverse and realistic data are the primary strengths of crowdsourcing \cite{rea2020crowdsourcing, vaughan2017making}. LLMs like GPT and BARD are trained on conventional text data from web and book corpora with no known established mechanism to validate the information. This may lead to errors or misinformation, which should be avoided, especially in healthcare. Using crowdsourced data, researchers can more easily take care of the required quality control. Additionally, adaptability in crowdsourcing enables iterative data collection based on model performance tailored to the researcher’s need. For example, in the case of training a predictive healthcare language model, if one label performs poorly because of inferior data quality in the first iteration, the researcher can iteratively collect additional data to improve that specific label’s performance.  

However, crowdsourcing requires data quality control, which is often overlooked. For example, Geiger enlisted four types of crowd work: crowd creation (crowd creates content), solving (crowd seeks solution to a specific problem), rating (crowd creates Ratings and reviews), and processing (collective effort of "the crowd" in completing tasks) \cite{geiger2011crowdsourcing}. In the case of crowd creation, which involves content generation, like responding to specific questions or writing reviews and blog posts, the evaluation criteria are often vague and poorly defined \cite{geiger2011crowdsourcing}, resulting in subpar data quality. Also, Allahbakhsh mentioned quality control in crowdsourcing as a critical issue \cite{allahbakhsh2013quality} in their work. This quality control challenge often results in inferior quality data, creating the loop of the "garbage in, garbage out" issue, as described by the father of modern computing, Charles Babbage \cite{babbage1864passages}. Hence, there is an increasing focus on the data-centric approach in modern artificial intelligence (AI) research, where data is methodically engineered as an essential step in creating reliable AI systems \cite{zha2023data}.

Following this data-centric approach, several researchers suggested various cures for low quality in crowdsourced data, broadly classified as pre-quality control schema (Pre QC) and post-quality control schema (Post QC). For instance, Allahbakhsh laid out the Pre QC schema among two essential dimensions: worker profile and task design \cite{allahbakhsh2013quality}. It mainly focuses on creating a suitable worker pool coupled with detailed task design at a granular level \cite{nilforoshan2017precog, allahbakhsh2013quality}. Reputation and credential (e.g., education, approval rate) based worker selection and defensive task design are essential parts of a Pre QC schema. In their work, Enrique showed that using adequate incentives (also a part of Pre QC) as an extrinsic motivator can make crowdsourcing successful \cite{estelles2018need}. However, Malone argued that adequate incentives could be solely due to monetary incentives harming the crowdsourcing effort \cite{malone2010collective}. Additionally, most of this Pre QC schema implementation is a hard-wired quality control technique typically embedded in the host website, and researchers cannot often customize it based on their specific requirements \cite{allahbakhsh2013quality}.

In the Post QC schema, the QC mechanism is applied once the data has been collected. Here, enhancement can be achieved through post-processing by implementing rigorous attention checks, setting quality thresholds for text data, and some generic text cleanup \cite{nilforoshan2017precog}. Besides that, expert review (domain expert assesses the data quality), majority consensus, contributor evaluation (evaluates a contribution based on the contributor's quality) \cite{allahbakhsh2013quality}, and workflow management \cite{kittur2011crowdforge, kulkarni2012collaboratively} are a few other ways Post QC can be implemented. However, it is acknowledged that Post QC can be resource-intensive, consuming valuable time and financial resources, thereby creating a significant gap in the quality control process.

To address this challenge, we developed a crowdsourcing (CS) framework integrating real-time quality control (real-time QC), acting as a discreet background filter that complements Pre QC efforts. Our framework strategically addresses the inherent data quality and integrity challenges associated with crowdsourced data, thus making it a more reliable and efficient solution. Through experimentation, expert, statistical, and ML evaluation, we demonstrate that crowdsourced data, as improved through our data quality control, can help increase the performance of LLMs, even when faced with limited resources.

%==============================
\section{Materials and Methods}\label{method}
%==============================

We collect text data to fine-tune a transformer-based language model (Bio-BERT \cite{lee2020biobert}) to predict signs and symptoms of autism from text data. We compare three stages for crowdsourcing quality control: Pre Quality Control (Pre QC), Real-time QC, and Post Quality Control (Post QC). We work with Amazon Mechanical Turk (AMT) and Qualtrics surveys.

\subsection{Pre QC}
During the pre-QC phase, we use a qualification task that requires AMT workers who reside in the United States, with a minimum work approval rate of 98\%. These AMT minimum requirements are a common
approach and not domain specific. The qualification task focuses on creating additional filters for the workers. Within the qualification task, we asked for demographic information and inquired about any personal or familial connections with autism or autism-related concerns, i.e., we posed 29 questions related to signs and symptoms of autism. While ATM offers a wide array of default qualifications, we customize qualifications to align with our domain specific objectives using the responses on the qualification task: AMT masters ( MM)  qualification for those with AMT a master’s degree, healthcare (HC) qualification for workers who are in the healthcare profession, graduate degree (GD) qualification for people with a minimum US graduate degree, and education (ED) for participants who are education professionals.

\subsection{Real-time QC}

The real-time Quality Control (publicly available on GitHub \footnote{https://github.com/porosantabarai/Crowdsourcing-Real-time-QC.git}) phase is executed simultaneously with the submission of responses by workers. Its primary objective is to identify and mitigate copy-and-paste answers. Workers generate responses once the Human Intelligence Task (HIT) is live on AMT. Then, we detect coherent text and nonsensical gibberish by HTML’s built-in spellcheck feature. For example, if a respondent typed nonsensical words during the Google search for that specific response, the search engine will return a blank page and a null value for the retrieved text vector. Using that information, we trigger an alert message to re-enter the response.

When coherent text is identified, we proceed to initiate an XHR API call, leveraging the Google Custom Search Engine API. Following the API call, the framework extracts and retrieves the surface text from the Google search results. Subsequently, it computes n-gram representations for the input text and the retrieved content. If at least one congruent n-gram is detected during this process, the response is assumed to be copied and fails the automated evaluation. For example, say text 1 (\textbf{A dietary restriction} implies restrictions on specific foods that an individual cannot or \textbf{will not consume}.) is the response text, and text 2 (\textbf{A dietary restriction} means restrictions on certain foods that a person \textbf{will not consume}.) is the retrieved text, then this will be recognized as copied response due to 2 bolded 3-gram matches. We provide an alert message, prompting the user to re-enter their response. This evaluation process is executed behind the scenes with little to no obtrusion to the worker. Our framework can also keep the metadata on the number of attempts before a successful submission for any necessary post-hoc analysis.

\subsection{Post QC}

After data collection, we carry out the Post QC of the data. Our primary goal in this phase is to detect and eliminate low-quality responses. One or more strategies can be used, such as defining strict performance cutoff criteria on attention and quality check questions, evaluating response completion times, and using human evaluations. For instance, a respondent who repeats an authentic response, approved by Real-time QC, for multiple questions unrelated to the specific symptom the question is asking can be eliminated in this phase of QC. We used multiple expert assessments to evaluate the applicability and quality of the collected data. 

\section{Study Design}

We compare the three types of quality control using four sequential studies. First, Experiment 1, or the "Pre QC" study, investigates the influence of pre-QC measures on the data. Then, the subsequent three experiments incorporate real-time QC features using the best quality control setting from Experiment 1. Experiment 2, denoted as "Paste," introduces a copy-paste restriction for respondents, preventing them from copying and pasting responses. Experiment 3, "GS," integrates custom Google search features (as outlined in the Real-time QC section). Finally, Experiment 4, "Paste+GS," combines the features from experiments two and three. We also collected baseline data (No QC) without custom qualification or CS implementation.

\subsection{Procedure}

We collected all the data, including four experimental conditions and No QC on the AMT platform. In the Pre QC experiment, we first posted the qualification task, which collected data regarding respondent demographics and their connections to individuals diagnosed with or displaying symptoms of autism. 

Then, a HIT comprising one of the 29 questions related to the signs and symptoms of autism was made available. For instance, "Describe an interaction with you or others where this person uses sounds that you don’t expect in a typical interaction"  is an example question related to the A1 symptoms prepared by pediatricians. After completing our qualification tasks, AMT workers could complete up to 29 HITs (all questions). In all experimental conditions, we requested ten responses per question on AMT. Workers were compensated with \$0.10 for completing the voluntary qualification task and \$0.40 for each question answered. There was an average delay of 12 days between the experiments.

\subsection{Evaluation}

We conduct our evaluation in the context of classifying text as ASD criteria. The American Psychiatric Association’s Diagnostic and Statistical Manual (DSM) outlines specific diagnostic criteria for autism spectrum disorder (ASD), consisting of 12 criteria classified into domains A, B, C, D, and E. Domain A focuses on "Persistent Deficits in Social Communication and Social Interaction" (A1-A3), while Domain B addresses "Restricted, repetitive patterns of behavior, interests, or activities" (B1-B4). To receive an ASD diagnosis per DSM-5 guidelines, an individual must meet three criteria from Domain A and at least two criteria from Domain B. We used Bio-BERT \cite{lee2020biobert} to label sentences in EHR with A1-A3 and B1-B4 labels. Our methodology involved retraining this model (Bio-BERT) on a dataset of 34,313 sentences extracted from 150 CDC-trained clinician reports. These reports were generated as part of the Autism Developmental Disabilities Monitoring (ADDM) surveillance by the Centers for Disease Control (CDC). Among 34,313 sentences, only 3,570 were labeled. We use 10-fold cross-validation.  We report precision and recall for the baseline model Bio-BERT (no data augmentation) and Experiments 1 to 4. Our goal is to augment this dataset with the newly collected data.

\subsubsection{Basic Quality Evaluation}

Two independent evaluators blindly assessed the overall data quality of the four experimental datasets. They categorized responses as "Overall Good" or "Overall Bad." Responses were marked as "Overall Good" if they were relevant and/or contained example behavior, while responses were marked as "Overall Bad" if irrelevant, clearly copied from another source, incoherent, or if the example behavior pertained to a different question. For example, this response ("The autistic person that I know groans a lot in conversations, even when the topic is lighthearted and does not warrant any annoyance or anything that should normally elicit a groan.") was given to a question related to A1 but actually it aligns with DSM criteria B1 hence marked as "Overall Bad. All the ratings were binary (1 or 0). Using the rating of the response from two evaluators, we calculated Cohen’s kappa (quantitative measure of agreement between two raters) for the “Overall Good” metric. We also perform a one-way ANOVA test among the ratings from two evaluators to check if there is any significant difference between the two evaluators.

\subsubsection{Domain Expert Evaluation}

A domain expert evaluated a random subset of the 175 responses from the Post QC data set. All examples were in random order. Five categories were used for this evaluation: Typical, Not Typical, Normal, EHR Match, and Exact Match, which assess response closeness to autism behavior. The first, Typical, assesses how well the behavior aligns with typical characteristics; not Typical asses atypical behavior but not commonly seen in autism, and Normal assesses the regular behavior; EHR Match assesses if the response behavior is typically encountered in HER. Finally, Exact Match assesses if the provided response matches exactly with the actual labels. Expert evaluated the first four categories on a five-point Likert scale from strong relevance (5) to o relevance (1) for the respective category and “Exact Match” as yes or no.

%==============================
\section{Results}\label{results}
%==============================
\subsection{Crowdsourced Dataset}

A total of 1200 workers completed our qualification tasks, and upon inspection, 982 were assigned four distinct qualifications (MM, HC, GD, and ED): 264 MM workers, 307 HC workers, 318 GD workers, and 93 ED workers. Subsequently, we published the HITs containing the 29 questions exclusively for the selected four cohorts (MM, HC, GD, and ED) to observe the effect of custom qualification on data quality. We requested ten random responses per question (implemented by AMT), resulting in 29*10*4 = 1,160 requested responses.

We selected the highest-performing qualification group from the Pre QC experiments for the subsequent three experiments. The GD group was the group with the highest quality data (see Table \ref{tab:humaneltn}). Following HITs were made accessible solely to this pool of 318 GD workers. Table \ref{tab:data} shows training data size in different experimental conditions by each DSM criteria. Our Post QC dataset encompasses 866 evaluated responses from the four experimental datasets. Additionally, we accumulated a baseline dataset of 1551 sentences for the same set of questions, labeled "NO QC," where no qualification controls were implemented, serving as a valuable point of comparison.

% \usepackage{booktabs}
% \usepackage{graphicx}
\begin{table}[ht]
\centering
\caption{Number of training sentences by different DSM criteria for all experimental conditions.}
\label{tab:data}
\resizebox{\columnwidth}{!}{%
\begin{tabular}{@{}cccccccc@{}}
\toprule
\textbf{DSM   Criterion} & \textbf{Bio-BERT Training} & \textbf{No QC} & \textbf{\begin{tabular}[c]{@{}c@{}}Pre QC\\ Experiment-1\end{tabular}} & \multicolumn{3}{c}{\textbf{Real-time QC}} & \textbf{Post QC} \\ \midrule
 &  &  &  & \textbf{Experiment-2: Paste} & \textbf{Experiment-3: GS} & \textbf{Experiment-4: Paste+GS} &  \\ \cmidrule(lr){5-7}
A1 & 855 & 200 & 192 & 50 & 50 & 50 & 124 \\
A2 & 471 & 216 & 129 & 30 & 30 & 30 & 83 \\
A3 & 524 & 204 & 160 & 40 & 40 & 40 & 91 \\
B1 & 539 & 228 & 160 & 40 & 40 & 40 & 144 \\
B2 & 338 & 203 & 160 & 40 & 40 & 40 & 92 \\
B3 & 146 & 215 & 119 & 30 & 30 & 30 & 124 \\
B4 & 697 & 284 & 240 & 60 & 60 & 60 & 208 \\ \midrule
\textbf{Total} & 3570 & 1550 & 1160 & 290 & 290 & 290 & 866 \\ \bottomrule
\end{tabular}%
}
\end{table}

%==============================
\subsection{Participants and Demographics}\label{llm}
%==============================
The demographic characteristics across the five experimental conditions reveal several variations (see Table \ref{tab:demog}). Females were more predominant in Experiment 4, "Paste+GS"  (54\%), compared to the other conditions, while males were more prominent across the other experimental conditions, with the highest being 63\% in Experiment 3. When considering race, White individuals dominated all conditions, especially in Experiment 1 (91.7\%) and Experiment 4 (76.9\%). Regarding ethnicity, non-Hispanic individuals were the majority across all experimental conditions, with higher percentages in Experiment 3 (96.3\%) and 4 (96.1\%). Bachelor's degree was the most prominent in all conditions but notably lower in Experiment 2 (38.2\%). Lastly, the age characteristics displayed minor variations among experiments, with the mean age ranging from 38 to 46.9 years, but no significant distinctions.

% \usepackage{booktabs}
% \usepackage{multirow}
% \usepackage{graphicx}
\begin{table}[]
\centering
\caption{Participants' characteristics on four and no experimental conditions {[}\% (count){]}. }
\label{tab:demog}
\resizebox{\columnwidth}{!}{%
\begin{tabular}{@{}llccccc@{}}
\toprule
\textbf{Variable} & \textbf{Choice} & \textbf{No QC (n=160) \tablefootnote{Education information is not available for No QC data because this property was collected as a part of Pre QC experiment.}} & \textbf{\begin{tabular}[c]{@{}c@{}}Pre QC\\ Experiment-1\\ (n=168)\end{tabular}} & \multicolumn{3}{c}{\textbf{Real-time QC}} \\ \midrule
\textbf{} & \textbf{} & \textbf{} & \textbf{} & \textbf{\begin{tabular}[c]{@{}c@{}}Experiment-2: Paste\\ (n=34)\end{tabular}} & \textbf{\begin{tabular}[c]{@{}c@{}}Experiment-3: GS\\ (n=54)\end{tabular}} & \textbf{\begin{tabular}[c]{@{}c@{}}Experiment-4: Paste+GS\\ (n=26)\end{tabular}} \\ \cmidrule(l){5-7} 
\multirow{2}{*}{Sex} & Female & 38.1 (61) & 37.5 (63) & 47.1 (16) & 37.0 (20) & 53.9 (14) \\
 & Male & 61.9 (99) & 62.5 (105) & 52.9 (18) & 63.0 (34) & 46.2 (12) \\ \midrule
\multirow{4}{*}{Race} & American Indian or Alaska Native & 1.9 (3) & 0.6 (1) & 0.0 (0) & 0.0 (0) & 0.0 (0) \\
 & Asian & 13.1 (21) & 4.2 (7) & 5.9 (2) & 3.7 (2) & 7.7 (2) \\
 & Black or African-American & 5 (8) & 1.8 (3) & 2.9 (1) & 1.9 (1) & 11.5 (3) \\
 & White & 80 (128) & 91.7 (154) & 91.2 (31) & 94.4 (51) & 76.9 (20) \\ \midrule
\multirow{2}{*}{Ethnicity} & Hispanic or Latino & 15 (24) & 18.5 (31) & 5.9 (2) & 3.7 (2) & 3.9 (1) \\
 & Not Hispanic or Latino & 85 (136) & 81.5 (137) & 94.1 (32) & 96.3 (52) & 96.1 (25) \\ \midrule
\multirow{6}{*}{Education} & Less than high school degree & NA & 0.0 (0) & 0.0 (0) & 0.0 (0) & 0.0 (0) \\
 & High school diploma & NA & 3 (5) & 0.0 (0) & 3.7 (2) & 11.5 (3) \\
 & Associate degree & NA & 1.2 (2) & 0.0 (0) & 1.9 (1) & 11.5 (3) \\
 & Bachelor's degree & NA & 53.6 (90) & 38.2 (13) & 57.4 (31) & 46.2 (12) \\
 & Master's degree & NA & 41.7 (70) & 61.8 (21) & 37.0 (20) & 30.8 (8) \\
 & Doctoral degree (PhD, MD,...) & NA & 0.6 (1) & 0.6 (1) & 0.0 (0) & 0.0 (0) \\ \midrule
\multirow{3}{*}{Age} & Mean & 39.8 & 38.8 & 46.9 & 38.0 & 42.5 \\
 & SD & 13.7 & 12.1 & 13.1 & 11.5 & 13.4 \\
 & Range & 22-70 & 21-70 & 27-70 & 25-70 & 23-70 \\ \bottomrule
\end{tabular}%
}
\end{table}

%==============================
\subsection{Basic Quality Evaluation}\label{Res_expert}
%==============================
Agreement between two evaluators was high (see Table \ref{tab:kappa-anova}), so we merged ratings and reported the average value. Highlighted rows from Table \ref{tab:humaneltn} show the quality of Pre QC data for four qualifications (MM, GD, HC, and ED). The GD cohort’s data showed the best data quality (68\%), while the ED cohort showed the highest percentage of data that evaluators classified as "Overall Bad" (70\%).  In the HC group, evaluators marked 20\% as bad. A substantial amount of the data remained unusable even with pre-quality control in place, highlighting the possible need for Real-time QC.

% \usepackage{booktabs}
% \usepackage{multirow}
% \usepackage{graphicx}
\begin{table}[ht]
\centering
\caption{Average Percentage Values of Measures Overall Good, Overall Bad, and Duplicate Across Four Experimental Conditions}
\label{tab:humaneltn}
\resizebox{\columnwidth}{!}{%
\begin{tabular}{@{}llccccc@{}}
\toprule
\textbf{Experiment} & \textbf{Description} & \multicolumn{1}{l}{\textbf{Qualification}} & \multicolumn{1}{l}{\textbf{Total Count (N)}} & \multicolumn{1}{l}{\textbf{Overall Good (\%)}} & \multicolumn{1}{l}{\textbf{Overall Bad (\%)}} & \multicolumn{1}{l}{\textbf{Duplicate (\%)}} \\ \midrule
\rowcolor[HTML]{C0C0C0} 
\cellcolor[HTML]{C0C0C0} & \cellcolor[HTML]{C0C0C0} & ED & 290 & 29.65 & 70.35 & 53.75 \\
\rowcolor[HTML]{C0C0C0} 
\cellcolor[HTML]{C0C0C0} & \cellcolor[HTML]{C0C0C0} & GD & 290 & 68.30 & 31.70 & 52.20 \\
\rowcolor[HTML]{C0C0C0} 
\cellcolor[HTML]{C0C0C0} & \cellcolor[HTML]{C0C0C0} & HC & 290 & 34.65 & 65.35 & 18.90 \\
\rowcolor[HTML]{C0C0C0} 
\multirow{-4}{*}{\cellcolor[HTML]{C0C0C0}1} & \multirow{-4}{*}{\cellcolor[HTML]{C0C0C0}Pre QC} & MM & 290 & 45.00 & 55.00 & 36.40 \\ \midrule
2 & Paste & GD & 290 & 36.55 & 63.45 &  \\
3 & GS & GD & 290 & 53.65 & 46.35 &  \\
4 & Paste+ GS & GD & 290 & 63.10 & 37.05 &  \\ \bottomrule
\end{tabular}%
}
\end{table}

Compared to only Pre QC data, we notice a consistent improvement in the data quality (approximately 20\%). Data collected with experimental conditions 2 and 3 combined achieved the highest quality (63\%). Comparing experiment 2 (Paste) with experiment 3 (GS), we see a 12\% to 22\% improvement in overall data quality. Overall, the "Combined" condition, which merged "Paste Restriction" and "Google Search," demonstrated both improved data quality ratings, underscoring the potential benefits of implementing Real-time QC in crowdsourcing. 

% \usepackage{booktabs}
% \usepackage{graphicx}
\begin{table}[ht]
\centering
\caption{Cohen’s Kappa and one-way ANOVA results for Overall Good across four experimental conditions.}
\label{tab:kappa-anova}
%\resizebox{\columnwidth}{!}{%
\begin{tabular}{@{}clll@{}}
\toprule
\multicolumn{1}{l}{\textbf{Experiment}} & \multicolumn{1}{l}{\textbf{Description}} & \multicolumn{1}{l}{\textbf{Cohen's Kappa}} & \multicolumn{1}{l}{\textbf{F-Value/p-value}} \\ \midrule
1 & Pre QC & 0.95 & 1.57/ 0.210 \\
2 & Paste  & 0.87 & 2.41/ 0.121 \\
3 & GS & 0.94 & 0.56/ 0.454 \\
4 & Paste+GS & 0.95 & 0.27/ 0.606 \\ \bottomrule
\end{tabular}%
%}
\end{table}

\subsection{Domain Expert Evaluation}

Table \ref{tab:clinician} shows the expert evaluation result of 175 randomly selected responses from  Experiment 4 (Paste=GS)  data. The result shows that 55\% (96) responses were intelligible, and 45\% (79) were unintelligible. Among 96 intelligible responses, the expert found that 75\% (72) responses exactly matched relevant diagnostic criteria. The rest of the 12 responses described multiple behaviors or other behaviors compared to the question asked. Among intelligible responses, we see a high agreement (average Likert scale rating of 4.4) of the expert on crowdsourced responses’ alignment with clinical EHR reports, indicating better data quality. Agreement on crowdsourced data describing normal behavior rather than autism is low, with an average Likert scale rating of 2.12. A higher Likert scale rating on Typical and EHR Match indicates better quality crowdsourced data.

% \usepackage{booktabs}
% \usepackage{graphicx}
\begin{table}[ht]
\centering
\caption{Expert evaluation of a random subset of the Experiment-4 data.}
\label{tab:clinician}
\resizebox{\columnwidth}{!}{%
\begin{tabular}{@{}ccccccccc@{}}
\toprule
\textbf{DSM Criterion} & \textbf{Counts} & \textbf{Unintelligible} & \textbf{Intelligible} & \textbf{Exact Match} & \textbf{Typical} & \textbf{Normal} & \textbf{Not Typical} & \textbf{EHR Match} \\ \midrule
A1 & 25 & 9 & 16 & 9 & 4.31 & 1.94 & 1.47 & 4.19 \\
A2 & 25 & 13 & 12 & 7 & 4.33 & 1.17 & 1.67 & 5.00 \\
A3 & 25 & 13 & 12 & 10 & 4.33 & 2.08 & 1.00 & 4.08 \\
B1 & 25 & 8 & 17 & 10 & 4.29 & 1.59 & 1.88 & 4.56 \\
B2 & 25 & 10 & 15 & 14 & 4.20 & 2.00 & 1.00 & 4.13 \\
B3 & 25 & 13 & 12 & 11 & 4.33 & 1.67 & 1.75 & 4.42 \\
B4 & 25 & 13 & 12 & 11 & 1.33 & 4.42 & 1.33 & 4.42 \\ \midrule
\textbf{Total} & 175 & 79 & 96 & 72 & NA & NA & NA & NA \\ \midrule
\textbf{Average} & NA & NA & NA & NA & 3.88 & 2.12 & 1.44 & 4.40 \\ \bottomrule
\end{tabular}%
}
\end{table}
%==============================
\subsection{LLM Evaluation}\label{Res_llm}
%==============================

For the LLM evaluation, we used data from four experiments and tested the impact of adding this to the baseline Bio-BERT data. Table 6 shows precision and recall. The table shows that transitioning from baseline (Bio-BERT) to "No-QC" reduced the model performance for precision but increased recall:  an average 23\% drop in precision and an average 13\% increase in recall.

There are notable differences between the different types of quality control. Moving from "No QC" to Experiment 1("Pre QC") generally results in improved precision across most labels, ranging from 3\% to 21\% improvement. On the other hand, the model’s recall value has dropped by as much as 10\%, indicating a higher chance of false negatives. Within the "Real-time QC" phase, different experiments ("Paste," "GS," and "Paste+GS") exhibit improved precision in most cases and, in general, similar recall values. For models with little baseline data, such as (B2 and B3), we see a gradual improvement in precision, with the highest (B2: 0.6471, B3: 0.6316) being in Experiment 4 data. The "Post QC" phase generally maintains precision at levels slightly better than "Pre QC." However, compared to Real-time QC, the performance gain of Post QC in terms of precision is 2\% to 4\%, with some labels having slightly inferior performance. Regarding the recall, most of the recall values decreased, ranging from a 3\% to 12\% reduction. Therefore, we can conclude that with proper real-time QC implemented, adding another layer of manual labor (in post-QC) did not help improve model performance a lot.

% \usepackage{booktabs}
% \usepackage{multirow}
% \usepackage{graphicx}
\begin{table}[ht]
\centering
\caption{Precision and recall measure of the Bio-BERT model on different experimental data.}
\label{tab:precrecall}
\resizebox{\columnwidth}{!}{%
\begin{tabular}{@{}c
>{\columncolor[HTML]{EFEFEF}}c 
>{\columncolor[HTML]{EFEFEF}}c cc
>{\columncolor[HTML]{EFEFEF}}c 
>{\columncolor[HTML]{EFEFEF}}c cccccc
>{\columncolor[HTML]{EFEFEF}}c 
>{\columncolor[HTML]{EFEFEF}}c @{}}
\toprule
 & \multicolumn{2}{c}{\cellcolor[HTML]{EFEFEF}} & \multicolumn{2}{c}{} & \multicolumn{2}{c}{\cellcolor[HTML]{EFEFEF}} & \multicolumn{6}{c}{\textbf{Real-time QC}} & \multicolumn{2}{c}{\cellcolor[HTML]{EFEFEF}} \\ \cmidrule(lr){8-13}
\multirow{-2}{*}{\textbf{DSM Criterion}} & \multicolumn{2}{c}{\multirow{-2}{*}{\cellcolor[HTML]{EFEFEF}\textbf{\begin{tabular}[c]{@{}c@{}}Bio-Bert \\ (Baseline)\end{tabular}}}} & \multicolumn{2}{c}{\multirow{-2}{*}{\textbf{No QC}}} & \multicolumn{2}{c}{\multirow{-2}{*}{\cellcolor[HTML]{EFEFEF}\textbf{\begin{tabular}[c]{@{}c@{}}Pre QC\\ Experiment-1\end{tabular}}}} & \multicolumn{2}{c}{\textbf{\begin{tabular}[c]{@{}c@{}}Experiment-2:\\ Paste\end{tabular}}} & \multicolumn{2}{c}{\textbf{\begin{tabular}[c]{@{}c@{}}Experiment-3:\\ GS\end{tabular}}} & \multicolumn{2}{c}{\textbf{\begin{tabular}[c]{@{}c@{}}Experiment-4:\\ Paste+GS\end{tabular}}} & \multicolumn{2}{c}{\multirow{-2}{*}{\cellcolor[HTML]{EFEFEF}\textbf{Post QC}}} \\ \midrule
 & \textbf{Precision} & \textbf{Recall} & \textbf{Precision} & \textbf{Recall} & \textbf{Precision} & \textbf{Recall} & \textbf{Precision} & \textbf{Recall} & \textbf{Precision} & \textbf{Recall} & \textbf{Precision} & \textbf{Recall} & \textbf{Precision} & \textbf{Recall} \\
\textbf{A1} & 0.7338 & 0.3269 & 0.5363 & 0.452 & 0.4758 & 0.5994 & 0.6218 & 0.4584 & 0.6234 & 0.4456 & 0.6267 & 0.452 & 0.5654 & 0.5129 \\
\textbf{A2} & 0.83 & 0.5929 & 0.5856 & 0.6929 & 0.785 & 0.6 & 0.7579 & 0.6929 & 0.7742 & 0.6858 & 0.7616 & 0.7072 & 0.804 & 0.5858 \\
\textbf{A3} & 0.7379 & 0.5 & 0.4046 & 0.6823 & 0.5107 & 0.6636 & 0.572 & 0.687 & 0.5868 & 0.6636 & 0.5917 & 0.6636 & 0.5488 & 0.6309 \\
\textbf{B1} & 0.6438 & 0.6732 & 0.4793 & 0.7648 & 0.5591 & 0.6798 & 0.5498 & 0.7582 & 0.565 & 0.7386 & 0.5693 & 0.7255 & 0.6122 & 0.6602 \\
\textbf{B2} & 0.7615 & 0.5646 & 0.4661 & 0.7279 & 0.5898 & 0.7143 & 0.6491 & 0.6667 & 0.6173 & 0.6803 & 0.6471 & 0.6735 & 0.6716 & 0.6259 \\
\textbf{B3} & 0.6512 & 0.35 & 0.44 & 0.5125 & 0.625 & 0.5 & 0.6316 & 0.45 & 0.6207 & 0.45 & 0.6316 & 0.45 & 0.5257 & 0.5125 \\
\textbf{B4} & 0.6425 & 0.6455 & 0.45 & 0.7591 & 0.5221 & 0.75 & 0.5257 & 0.791 & 0.5134 & 0.7864 & 0.4958 & 0.8046 & 0.5289 & 0.75 \\ \midrule
\textbf{Average} & 0.7144 & 0.5219 & 0.4803 & 0.6559 & 0.5811 & 0.6439 & 0.6154 & 0.6435 & 0.6144 & 0.6358 & 0.6177 & 0.6395 & 0.6081 & 0.6112 \\ \bottomrule
\end{tabular}%
}
\end{table}

\subsection{Replication Evaluation of the Proposed Mechanism}

A comprehensive evaluation study was conducted to evaluate the effectiveness of the mechanisms designed to detect copied and Google-searched (GS) responses. Using the same set of 29 questions employed for data collection, new responses were collected and categorized into three groups: "Authentic," obtained from parental data collected through Qualtrics and verified by the human expert; "Copied," generated by performing a Google search for each question; and "Paraphrased," derived from the Copied responses using the ChatGPT paraphrasing mechanism. Subsequently, all responses were subjected to our detection tool to assess its robustness. Table \ref{tab:evltn} shows that the tool detected all 29 Authentic and 29 Copied responses. However, in the case of Paraphrased responses, the tool detected 11 out of 29, leaving 18 undetected instances.

% \usepackage{booktabs}
\begin{table}[ht]
\centering
\caption{Robustness assessment of the proposed mechanism.}
\label{tab:evltn}
\begin{tabular}{@{}lccc@{}}
\toprule
 & \multicolumn{1}{l}{\textbf{Authentic}} & \multicolumn{1}{l}{\textbf{Copied}} & \multicolumn{1}{l}{\textbf{Paraphrased}} \\ \midrule
\textbf{Detected} & 29 & 29 & 11 \\
\textbf{Undetected} & 0 & 0 & 18 \\ \midrule
\textbf{Total} & 29 & 29 & 29 \\ \bottomrule
\end{tabular}
\end{table}
%==============================
\section{Discussion and Conclusion}\label{discussion}
%==============================

We provided a comprehensive QC outline for improving the performance of LLMs in healthcare through the strategic use of crowdsourced data and quality control measures. We observed significantly improved recall in models with crowdsourced data filtered by data control compared to the Bio-BERT baseline model trained on EHR data, but precision dropped. However, improved precision and minor improvement in recall were observed in four experiments compared to no QC-implemented model. Experimenting with a larger data set is needed to investigate this issue further. We showed that only Pre QC is insufficient, and Post QC might be redundant and resource-intensive while Real-time QC is in place. In this study, we experimented with data quality and how LLM’s performance can be improved with crowdsourcing. In forthcoming studies, we plan to juxtapose synthetic data augmentation and resampling techniques while also addressing the current limitation in paraphrased response detection by integrating a semantic understanding mechanism between responses. 

%==============================
\section{Acknowledgements}\label{acknowledgements}
%==============================
This project was supported in part by grant number R01MH124935 from the National Institute of Mental Health. Part of the data presented was collected by the Centers for Disease Control (CDC) and Prevention Autism and Developmental Disabilities Monitoring (ADDM) Network supported by CDC Cooperative Agreement Number 5UR3/DD000680 and by the University of Arizona FY23 BIO5 Rapid Grant.
%==============================
\bibliographystyle{vancouver}
\bibliography{literature}
%==============================
\end{document}